\documentclass[sigconf]{aamas} 

\usepackage{balance} 

\usepackage{graphicx}
\usepackage[medium,compact]{titlesec}

\usepackage{tikz}
\usepackage{comment}
\usepackage{amsmath} 
\usepackage{color}
\usepackage{amsmath}

\usepackage{booktabs}
\usepackage[utf8]{inputenc} 
\usepackage[T1]{fontenc}    
\usepackage{url}            
\usepackage{booktabs}       
\usepackage{amsfonts}       
\usepackage{nicefrac}       
\usepackage{microtype}      
\usepackage{xcolor}         
\usepackage{subfigure}
\usepackage[bb=boondox]{mathalfa}

\usepackage{multirow}
\usepackage{algorithmic}
\usepackage{algorithm}
\usepackage{rotating} 

\usepackage[capitalize]{cleveref}

\usepackage[utf8]{inputenc} 
\usepackage[T1]{fontenc}    
\usepackage{hyperref}       
\usepackage{url}            
\usepackage{booktabs}       
\usepackage{amsfonts}       
\usepackage{nicefrac}       
\usepackage{microtype}      

\usepackage{float}

\usepackage{amsthm}
\usepackage{caption}

\newcommand{\name}{MAGE-X}

\setcopyright{ifaamas}
\acmConference[AAMAS '23]{Proc.\@ of the 22nd International Conference
on Autonomous Agents and Multiagent Systems (AAMAS 2023)}{May 29 -- June 2, 2023}
{London, United Kingdom}{A.~Ricci, W.~Yeoh, N.~Agmon, B.~An (eds.)}
\copyrightyear{2023}
\acmYear{2023}
\acmDOI{}
\acmPrice{}
\acmISBN{}

\acmSubmissionID{442}

\title[AAMAS-2023 Formatting Instructions]{Learning Graph-Enhanced Commander-Executor \\for Multi-Agent Navigation}

\author{Xinyi Yang$^{1}$, Shiyu Huang$^{2}$, Yiwen Sun$^{3}$, Yuxiang Yang$^{1}$, Chao Yu$^{1,4}$, Wei-Wei Tu$^{2}$, \\Huazhong Yang$^{1\dag}$, Yu Wang$^{1\dag}$\\}
\affiliation{\institution{$^1$ Tsinghua University, $^2$ 4Paradigm Inc., $^3$ Fudan University, $^4$ Shanghai Artificial Intelligence Laboratory}\country{$^\dag$ Corresponding Author}}
\email{yang-xy20@mails.tsinghua.edu.cn}

\begin{abstract}
This paper investigates the multi-agent navigation problem, which requires multiple agents to reach the target goals in a limited time. Multi-agent reinforcement learning~(MARL) has shown promising results for solving this issue. 
However, it is inefficient for MARL to directly explore the (nearly) optimal policy in the large search space, which is exacerbated as the agent number increases~(e.g., 10+ agents) or the environment is more complex~(e.g., 3$D$ simulator). 
Goal-conditioned hierarchical reinforcement learning~(HRL) provides a promising direction to tackle this challenge by introducing a hierarchical structure to decompose the search space, where the low-level policy predicts primitive actions in the guidance of the goals derived from the high-level policy. 
In this paper, we propose \emph{\underline{M}ulti-\underline{A}gent \underline{G}raph-\underline{E}nhanced Commander-E\underline{X}ecutor}~({\name}), a graph-based goal-conditioned hierarchical method for multi-agent navigation tasks. {\name} comprises a high-level \emph{Goal Commander} and a low-level \emph{Action Executor}. The \emph{Goal Commander} predicts the probability distribution of the goals and leverages them to assign the most appropriate final target to each agent. 
The \emph{Action Executor} utilizes graph neural networks~(GNN) to construct a subgraph for each agent that only contains its crucial partners to improve cooperation. Additionally, the Goal Encoder in the \emph{Action Executor} captures the relationship between the agent and the designated goal to encourage the agent to reach the final target.
The results show that {\name} outperforms the state-of-the-art MARL baselines with a 100\% success rate with only 3 million training steps in multi-agent particle environments~(MPE) with 50 agents, and at least a 12\% higher success rate and 2$\times$ higher data efficiency in a more complicated quadrotor 3$D$ navigation task.

\end{abstract}


\newcommand{\BibTeX}{\rm B\kern-.05em{\sc i\kern-.025em b}\kern-.08em\TeX}

\begin{document}


\pagestyle{fancy}
\fancyhead{}


\maketitle 


\section{Introduction}
Navigation is a typical task in the intelligent agent system, applied in a wide range of applications, such as autonomous driving~\cite{autonomousdriving,autonomousdriving2}, logistics and transportation~\cite{logistics,logistics2}, and search and rescue for disasters~\cite{rescue,rescue2}. In this paper, we consider a multi-agent navigation task where multiple agents simultaneously move to the target goals in a cooperative fashion. Multi-agent reinforcement learning~(MARL) has attracted significant attention due to its powerful expressiveness in multi-agent navigation tasks~\cite{maans,multinavigation, multi-navigation, inferenceMultiNavigation}.

The common approach to searching for the near-optimal solution in MARL~\cite{mappo,mat} is to directly train a policy that produces environmental actions for agents. However, learning the strategy directly from large search space results in low data efficiency, which is more severe as the number of agents or the complexity of the environment increases. Therefore, the existing methods in navigation tasks target simple scenarios with few agents.
Goal-conditioned HRL~\cite{gchrl_1,gchrl_2,gchrl_3,gchrl_4,gchrl_5,gchrl_6,gchrl_7} has been recognized as an effective paradigm to address this problem, comprising a high-level policy that breaks the original task into a series of subgoals and a low-level policy that aims at the arrival of these subgoals. Recent works~\cite{maser,LGA} in goal-conditioned HRL mainly focus on developing high-level policies for providing agents with appropriate subgoals. However, designating the subgoals may confuse agents on which target goals they should reach.


Graph neural networks~(GNN) have been widely applied in multi-agent cooperative tasks due to their effective learning of agents' graph representations. The literature on GNN~\cite{MAGIC,DICG,gnn_3} in MARL constructs a graph whose nodes represent agents' information to model the interaction among them and encourage agents to cooperate. To enhance the cooperation among agents, we can leverage GNN in the low-level policy in HRL to capture the relationship of agents and express preferences for different teammates.



To improve data efficiency and cooperation, we propose \emph{\underline{M}ulti-\underline{A}gent \underline{G}raph-\underline{E}nhanced Commander-E\underline{X}ecutor}~({\name}), a graph-based goal-conditioned hierarchical framework in multi-agent navigation tasks. {\name} consists of two components, the high-level {\bf Goal Commander} and the low-level {\bf Action Executor}. For the target-goal assignment, the {\bf Goal Commander} infers the probability distribution of target goals to assign the most appropriate target goal for each agent instead of the subgoal.
As a result, the multi-agent navigation is converted to multiple single-agent navigation tasks in the multi-agent environment, in which each agent is required to reach the designated goal while avoiding collisions.
In the {\bf Action Executor}, we take advantage of GNN to perceive the relationship among agents and produce a subgraph for each agent to decide to whom they should pay attention.
After that, the State Extractor receives the correlation between the agent and its target goal from the Goal Encoder and the agent's embedded feature in the subgraph to empower the team representation with strong goal guidance, promoting agents to reach target goals. The suggested scheme is challenged against MARL baselines in multi-agent particle environments~(MPE)~\cite{mpe} with a massive number of agents and a more complicated quadrotor navigation task~\cite{drone}. The experimental results demonstrate that {\name} outperforms the MARL baselines, achieving a 100\% success rate with only 3 million training steps in MPE with 50 agents. Furthermore, {\name} attains at least a 12\% higher success rate and 2$\times$ higher data efficiency in the quadrotor navigation task.

Our contributions can be summarized as follows:
\setlength{\parskip}{0pt} 
\setlength{\itemsep}{0pt plus 1pt}
\begin{itemize}
\item We introduce a graph-based goal-conditioned framework in multi-agent navigation tasks, \emph{Multi-Agent Graph-enhanced Commander-Executor}~({\name}), to solve the problem of data efficiency and cooperation in large search space.
\item We develop a high-level Goal Commander, which utilizes the probability distribution of goals to allocate each agent to the most appropriate target goal.
\item We propose the low-level Action Executor, which adopts GNN to improve the coordination, and the Goal Encoder and the State Extractor to encourage agents to complete the task.
\item {\name} converges much faster and substantially outperforms MARL algorithms in multi-agent particle environments~(MPE)~\cite{mpe} with a massive number of agents and a quadrotor navigation task~\cite{drone}.
\end{itemize}

\section{Related Work}

%
\subsection{Navigation}
Navigation has been widely investigated in recent years, where RL has shown its ability to solve various applications~\cite{navigation1, navigation2, navigation3, save}. For example, Rao~\cite{visualnavigation} presents a model-embedded actor-critic architecture for the multi-goal visual navigation task. This embedded model consists of two auxiliary task modules, a path closed-loop detection module to understand whether the state is experienced and a state-target matching module to distinguish the difference between states and goals. Furthermore, Zhu~\cite{navigation1} introduces a target-driven actor-critic model to achieve greater adaptability and flexibility for the target-driven visual navigation task.

As for multi-agent navigation tasks in MARL~\cite{multinavigation, multi-navigation, inferenceMultiNavigation,maans}, the difficulties lie in data efficiency and cooperation in large space spaces, which is exacerbated as agent number increases and the environment becomes more complicated. EPS~\cite{multi-navigation} is introduced to enhance exploration efficiency and improve sample efficiency in multi-robot mapless navigation, which uses the evolutionary population periodically generated from robots’ policies to search for different and novel states. Furthermore, Xia~\cite{inferenceMultiNavigation} proposes an inference-based hierarchical reinforcement learning framework~(IHRL) to address the multi-agent cooperative navigation problem via the interplay of high-level inference and low-level actions. The proposed {\name} specializes in navigation tasks and outperforms the MARL baselines with high sample efficiency.


\subsection{Goal-conditioned HRL}
Goal-conditioned hierarchical reinforcement Learning~(HRL) has shown its capability in a wide range of tasks~\cite{goalconditioned1,HForesight,goalhierarchical} with a hierarchy consisting of high-level and low-level policies. The high-level policy generates intermediate subgoals every global timestep, which is regarded as a goal guidance of the low-level policy. 

Kreidieh~\cite{multiHRLgoal} addresses the challenges of the interactions between high-level and low-level agents by introducing inter-level cooperation. This inter-level cooperation is given by modifying the high-level policy's objective function and subsequent gradients. 
Another representative is LGA~\cite{LGA}, where the subgoal assignment is parameterized as latent variables to be trained. LGA directly provides primitive actions for agents depending on latent variables to accomplish multi-agent tasks. 
MASER~\cite{maser} automatically produces subgoals for agents from the experience replay buffer relying on both individual and total Q-values. Besides, it adopts the individual intrinsic reward for each agent to reach the assigned subgoals and maximize the joint action value. 
However, the generation of subgoals may be inefficient since the correspondence between subgoals and the final goals is implicit. In this paper, {\name} utilizes the goal-conditioned hierarchical framework in multi-agent navigation tasks to improve the sample efficiency, where the high-level policy deals with target-goal assignment and the low-level policy deals with the action execution for each agent.

\subsection{Graph Neural Networks}
Graph neural networks~(GNN)~\cite{gnn} are broadly used due to their effective learning of graph representations and the ability to capture the relationships of different graphs. Recently, several works have applied GNN in multi-agent systems to model agents' interactions. HAMA~\cite{gnn_hierarchical} proposes a hierarchical graph attention network that captures the underlying relationships at the agent-level and the group-level, enhancing generalization and scalability. MAGIC~\cite{MAGIC} is a novel graph communication protocol that implies the topology of agents' interactions, helping agents decide when to communicate and with whom to communicate. DICG~\cite{DICG} leverages dynamic coordination graphs to infer joint actions and values implicitly. GCS~\cite{GCS} learns coordinated behaviors by factorizing the joint team policy into a graph generator and a graph-based coordinated policy. The generator captures the underlying dynamic correlation of agents, which is then exploited by the graph-based coordinated policy. The Action Executor of {\name} benefits from GNN to capture the team representation and introduces a Goal Encoder and a State Extractor to strengthen the expression of target goals.

\section{Preliminary}

\begin{figure*}[t!]
	\centering
    \includegraphics[width=1.0\linewidth]{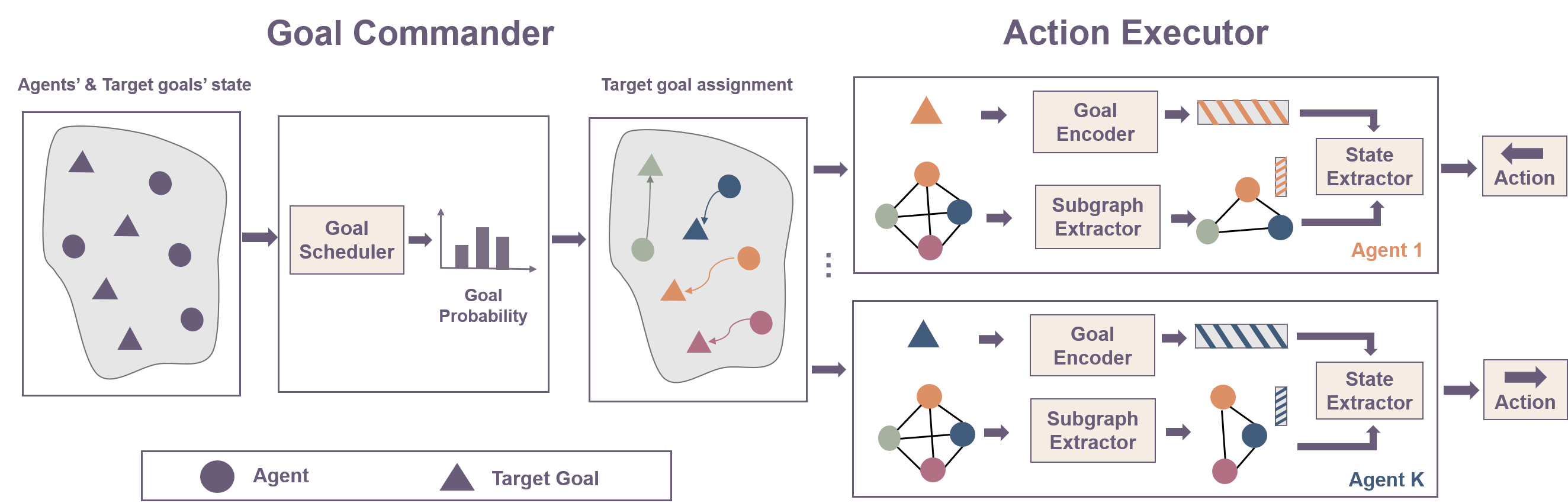}
	\centering \caption{{Overview of \emph{Multi-Agent Graph-enhanced Commander-Executor}}  (\name).}
\label{fig:Framework}
\end{figure*}
\subsection{Dec-MDPs}
In this paper, we consider a variant of MDP to solve the decentralized control problem in stochastic environments called Decentralized Markov Decision Processes~(Dec-MDPs).

Here, the multi-agent Dec-MDPs problem is formulated as:
\begin{equation}
<\mathcal{N},\mathcal{S}, \mathcal{A}, \mathcal{T}, R, G, \mathcal{O},\gamma>,
\end{equation}
where $\mathcal{N}\equiv\{1,...,n\}$ is a set of $N=|\mathcal{N}|$ agents. Note that $\mathcal{S}$ is a set of global states in the assumption that $\mathcal{S}$ is jointly observable. 
$\mathcal{A}$ is the action space of each agent, and $\mathbf{A}\equiv \mathcal{A}^N$ is the joint action space. $\mathcal{O}$ is the observation space. $R$ represents the reward function, and 
$R(s, \mathbf{a}, s')$ is the reward obtained from the transition of joint actions $\mathbf{a}\in\mathbf{A}$ from the state $s\in\mathcal{S}$ to the state $s^{\prime}\in\mathcal{S}$. $\gamma \in [0, 1)$ is the discount factor. $\mathcal{T}(s,\mathbf{a},s'):\mathcal{S}\times\mathbf{A}\times \mathcal{S}\mapsto [0,1]$ is the dynamics function denoting the transition probability. 
$G$ is the observation function, and $G(s, \mathbf{a}, s^{\prime}, o_i)$ is the probability of agents $i\in\mathcal{N}$ seeing observation $o_i\in\mathcal{O}$. 
Each agent has a policy $\pi^i(a^i_t|o^i_{1:t})$ to produce action $a^i_t$ from observations $o^i_{t}$ at step $t$. And agents need to maximize the expected discounted return $\mathbb{E} \sum^{\inf}_{l=0}\gamma^{l} r_{t+l}$, where $r_t=R(s_{t}, \mathbf{a}_t, s_{t+1})$ is the joint reward at step $t$.


\subsection{Graph Neural Networks}
Graph neural networks~(GNN) are a special type of neural network capable of dealing with data in the graph structure. The critical ingredient of GNN is pairwise message passing, i.e., graph nodes iteratively update their representations by exchanging information with their neighbors. The general formula is given as below:

\begin{equation}
    h_i^{(l)} = \sigma ( \sum_{j\in N_i} \frac{1}{\sqrt{d_i d_j}}(h_j^{l-1}W^{(l)}) ),
    \label{eq:GNN}
\end{equation}
where $h_i$ is the feature vector of node $i$. $N_i$ represents a set of neighbouring nodes of $i$ and $W^{(l)}$ is the learnable weights in the layer $l$. $\sigma$ is the activation function. $d_i$ is the dimension of feature stored in node $i$. Equation~\ref{eq:GNN} shows that the feature of node $i$ will be influenced by its neighbors. In this paper, we use graph convolutional networks~(GCN)~\cite{gcn} to model the interaction of agents. GCN is a variant of convolution neural networks~(CNN) to be applied in the data with graph structure.

\section{Methodology}

\begin{figure*}[t!]
	\centering
    \includegraphics[width=0.95\linewidth]{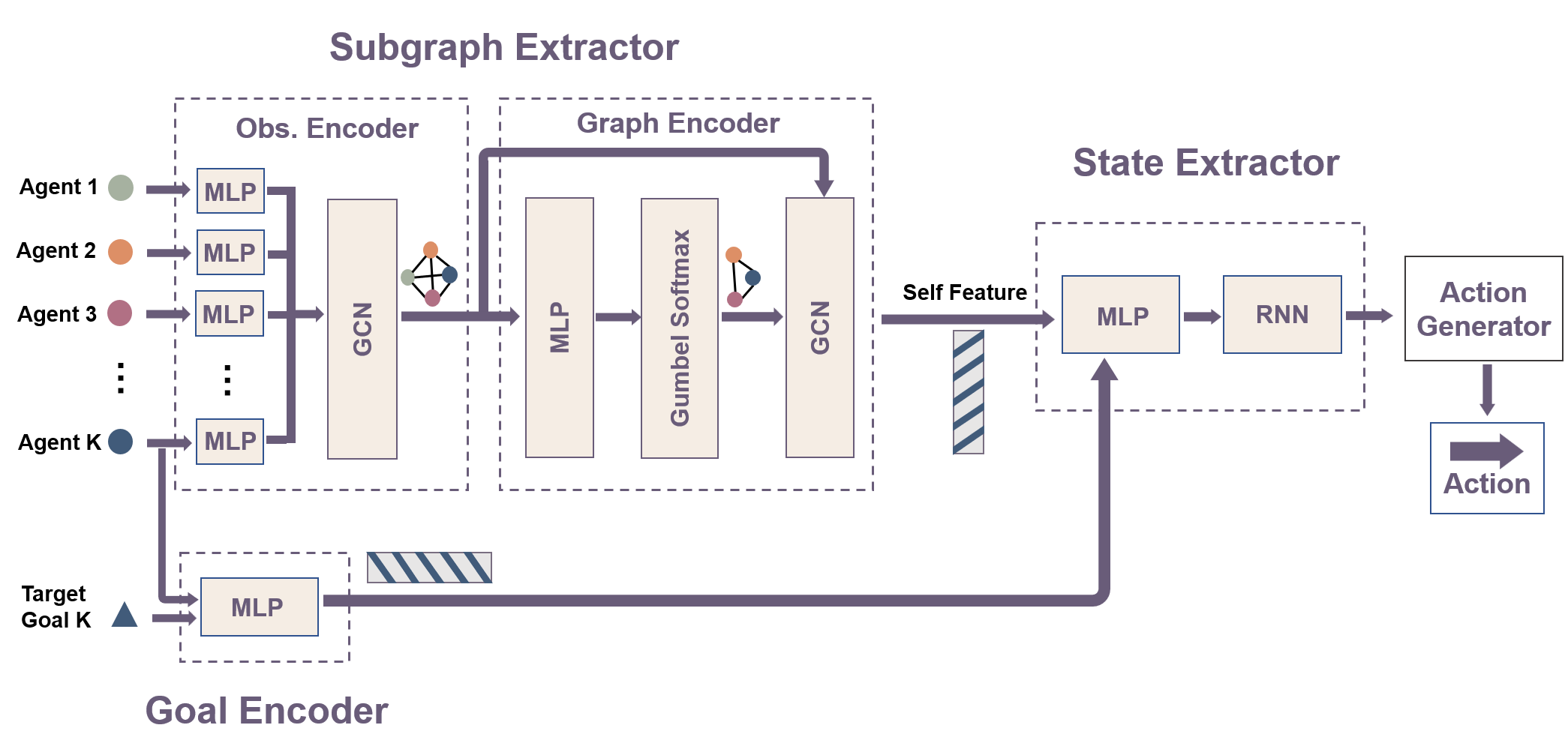}
	\centering \caption{Workflow of Action Executor, including a Subgraph Extractor, a Goal Encoder, a State Extractor for representation learning and an Action Generator.}
\label{fig:architecture}
\end{figure*}
In this section, we introduce the proposed framework, {\name}, to improve sample efficiency and the cooperation in multi-agent navigation tasks. The overview of our framework is demonstrated in ~\cref{fig:Framework}. {\name} comprises two components, the Goal Commander and the Action Executor. The high-level Goal Commander follows the principles of centralized-training-centralized-execution (CTCE). In contrast, the low-level Action Executor is in a decentralized setting with partial observation of $N$ agents, where agent $k$ receives local observation, $o_{t}^k$, at step $t$. Agent $k$ learns the policy, $\pi_k$, to produce a distribution over actions at each time step $t$, $a_{t}^k\sim\pi_{k}(a_{t}^k|o_{t}^k)$. The process of a navigation task begins with the Goal Scheduler in the Goal Commander receiving the spawn locations of all the agents and the target goals. The probability distribution of the target goals is predicted by the scheduler, and each agent is assigned the most appropriate target goal rather than the subgoal. Therefore, the multi-agent navigation is converted to multiple single-agent navigation tasks in the multi-agent environment, where each agent is required to reach the given goal as quickly as possible while avoiding collisions. Take agent $k$ as an example. The Subgraph Extractor takes in the observation of agent $k$ and other agents and produces the subgraph of agent $k$ only including its crucial neighbors to improve cooperation. The relationship of agent $k$ and its target goal is extracted from the Goal Encoder, which is then sent to the State Extractor combined with the feature of agent $k$ in the subgraph to endow the team representation with strong goal guidance. Finally, agent $k$ takes the preliminary action from the Action Generator to complete the navigation task. 
\subsection{Goal Commander}
Goal assignment is a long-studied maximal matching problem, especially in scenarios with large-scale agents. The Goal Commander builds upon a Goal Scheduler module for target-goal assignment, which produces the probability distribution of target goals. The designed reward is related to the distance cost of the Hungarian algorithm~\cite{Hungarian}, a state-of-the-art classical method to tackle the combinatorial optimization algorithm in graph theory.   

\textbf{Goal Scheduler}: This module is made up by a Multi-Layer Perception~(MLP) layer, $f_{sche}$, that takes the positions of all agents, $P_a$, and the positions of target goals, $P_g$, as input. We obtain the probability of the target goals, $P_{goal}$, by computing the softmax operator over the output of $f_{sche}$:
    \begin{equation}
    P_{goal} = Softmax(f_{sche}(P_a,P_g)).
    \end{equation}
Thereafter, $P_{goal}$ is ranked in decreasing order and the reordered goals $1...N$ are sequentially assigned to agent $1...N$. The reward of the Goal Scheduler, $R_{c}$, represents the distance cost of our assignment strategy, $C_{c}$, against the distance cost of the Hungarian algorithm, $C_{h}$:
    \begin{equation}
    R_{c} = 1-\frac{C_{c}}{C_{h}}.
    \end{equation}


\subsection{Action Executor}
The Action Executor is designed for high cooperation where agents speedily reach the designated goal with little collision in the multi-agent environment. The workflow of the Action Executor is illustrated in~\cref{fig:architecture}. It consists of the Subgraph Extractor, the Goal Encoder, the State Extractor, and the Action Generator. The Subgraph Extractor encodes the observations of all agents and yields a subgraph for agent $k$ only containing the crucial teammates. 
The Goal Encoder extracts the correlation of agent $k$ and its assigned goal, which is then fed into the State Extractor with the feature of agent $k$ in the subgraph to endow the team representation with target goal guidance. Finally, the Action Generator produces the action for the agent $k$. 

The environmental reward for each agent, $R_{e}$, is the linear combination of the complete bonus, $R_b$, the distance penalty, $R_d$, and the collision penalty, $R_c$:
 \begin{equation}
    R_{e} = \alpha{R_b}+\beta{R_d}+\gamma{R_c},
    \end{equation}
where $\alpha$, $\beta$ and $\gamma$ are the coefficient of $R_b$, $R_d$ and $R_c$, respectively.


\textbf{Subgraph Extractor: }The Subgraph Extractor is comprised of the Observation Encoder and the Graph Encoder. In the Observation Encoder, we apply an MLP layer $f_{o}$ to encode the observations of all agents and GCN to produce a fully connected graph of agents, $G_{a}$. This can be formulated as:
 \begin{equation}
G_a = GCN\left(\left(f_o(o^1),f_o(o^2),..f_o(o^N)\right),A_f\right),
    \end{equation}
where $A_f$ is denoted as an adjacent matrix of a fully connected graph. In the Graph Encoder, we compute the gumbel softmax~\cite{gumbel_softmax} over the feature of $G_{a}$ updated by an MLP layer, $f_{g}$. Thus, we can obtain an adjacent matrix of $A_s^k$ through it:
\begin{equation}
    A_s = Gumbel\_Softmax(f_g(G_a)).
    \end{equation}
    
Afterwards, we use GCN to generate $G_s^k$:
    \begin{equation}
    G_s^k = GCN(G_a,A_s^k).
    \end{equation}
The feature of the agent $k$ in $G_s^k$ is then sent to the State Extractor. Note that there are $l_{g}$ blocks in the Graph Encoder.

\textbf{Goal Encoder and State Extractor: }To capture the correlation between the agent $k$ and its assigned goal, the Goal Encoder comprises an MLP layer $f_{goal}$, and takes in the target goal's position $P_g^k$, and the agent's observation $o^k$. This can be formulated as:
\begin{equation}
    E_{goal}^{k} = f_{goal}(P_g^k,o^k).
    \end{equation}
    
Receiving the relationship between the agent and its target goal and the embedded feature of agent $k$ in $G_{s}$, $E_{a}^k$, we leverage the State Extractor to enhance the team representation with target goal guidance and output $E_{all}^k$. The State Extractor consists of an MLP layer $f_{state}$, and recurrent neural networks~(RNN) in consideration of the high correlation between current and historical states:
\begin{equation}
    E_{all}^k = RNN\left(f_{state}\left(concat\left(E_{goal}^k,E_{a}^k\right)\right)\right).
    \end{equation}
\subsection{Multi-agent Commander-Executor Training}


\begin{algorithm}
    \caption{Training Procedure of \emph{\name}}
    \label{algo:algo_mager}
    \begin{algorithmic}[1]
        \REQUIRE The positions $P_a$ of all agents, the joint observation $\mathbf{o}$ of all agents , and the positions $P_g$ of target goals.
        \ENSURE Final policy $\pi_\theta$ for the Goal Commander and $\pi_{\theta'}$ for the Action Executor.
        \STATE {\bf Initialize:} agents number $N$, maximal steps in an episode $T_e$, params in the Goal Commander $\theta$, $\phi$, params in the Action Executor $\theta'$, $\phi'$, the Goal Commander buffer $D$, the Action Executor buffer $D'$.
        \WHILE{$\theta$ and $\theta'$ not converges}
            \STATE{Reset environment and get $P_a$, $P_g$.}
            \STATE {\bf Initialize:} the step-count $t\leftarrow 1$ in $D$, the step-count $t'\leftarrow 1$ in $D'$.
            \STATE{ $P_{goal}\leftarrow$  \textbf{Goal\_Commander$(P_{a}$, $P_{g})$}.}
            \STATE{Calculate $\pi_\theta(a_t|\mathbf{o}_t)$ and $V_\phi(\mathbf{o}_t)$.}
            \STATE{Perform $a_{t}\sim\pi_\theta(a_t|\mathbf{o}_t)$.}
            \WHILE{$t'<T_e$ and not terminal}
                \STATE{$E_{all}^k\leftarrow$ \textbf{Action Executor ($\mathbf{o}$, $P_g^k$) } for each agent $k$}
                \STATE{Calculate $\pi_{\theta'}({a'^k_{t'}} |{o'^k_{t'}})$ and $V_{\phi'}({o'^k_{t'}})$ for each agent $k$.}
                \STATE{Perform $a'^k_{t'}\sim \pi_{\theta'}(a'^k_{t'} |{o'^k_{t'}} )$ for each agent $k$.}
                \STATE{Receive $r'_{t'}$ and ${o'^k_{t'+1}}$ for each agent $k$.} 
                \STATE{Store $({o'^k_{t'}},{a'^k_{t'}},\pi_{\theta'},({a'^k_{t'}}|o'_{t'}),V_{\phi'}({o'^k_{t'}}),r'_{t'},{o'}^k_{t'+1})$ in $D'$.}
                \STATE{$t'\leftarrow t'+1$}
            \ENDWHILE
            \STATE{Receive $r_t$ and $o_{t+1}$.}
            \STATE{$t\leftarrow t+1$}
            \STATE{Store $(o_t,a_t,\pi_\theta,(\mathbf{a}_t|\mathbf{o}_t),V_\phi(o_t),r_t,o_{t+1})$ in D.}
            \STATE{Perform update of $\theta$, $\phi$, $\theta'$ and $\phi'$.}
        \ENDWHILE
    \end{algorithmic}
\end{algorithm}

{\name}, following the goal-conditioned MARL framework, trains two policy networks for the Goal Commander and the Action Executor separately, which are optimized by maximizing the accumulated reward in the entire episode via reinforcement learning. We use Multi-agent Proximal Policy Optimization~(MAPPO)~\cite{mappo}, a multi-agent variant of Proximal Policy Optimization~(PPO)~\cite{ppo}, as the policy optimizer.

As shown in \cref{algo:algo_mager}, {\name} takes in the positions of agents, the observations of agents, and the positions of landmarks. Then, it produces the final policy $\pi_\theta$ for the Goal Commander and $\pi_{\theta'}$ for the Action Executor. First, we initialize the number of agents along with several training parameters, including maximal steps, parameters in the Goal Commander, parameters in the Action Executor, the Goal Commander reply buffer, $D$, and the Action Executor reply buffer, $D'$~(Line 1). 
The Goal Commander only performs one action every episode to assign the goals to agents in the beginning~(Line $5\sim7$) and receives the reward $r_t$ and the subsequent observation $o_{(t+1)}$ at the end of each episode, where $t$ represents the step-count in $D$~(Line 15 and 16). 
Thereafter, we update $\theta$ for the policy network $\pi_\theta$, and $\phi$ for the value network $V_{\phi}$, in the Goal Commander. 
Regarding the Action Executor, it outputs the action ${a'}^k_{t'}$ for each agent $k$, and stores a group of data in $D^{'}$ at each timestep $t'$~(Line $9\sim 13$). Similarly, we update $\theta'$ for the policy network $\pi_{\theta'}$, and $\phi'$ for the value network $V_{\phi'}$, in the Action Executor. \looseness=-1

\section{Experiments}

\subsection{Task Setup}
\label{sec:task_setup}
\begin{figure}[ht!]
\captionsetup{justification=centering}
	\centering
    \subfigure
	{\centering
        {\includegraphics[height=0.25\textwidth]{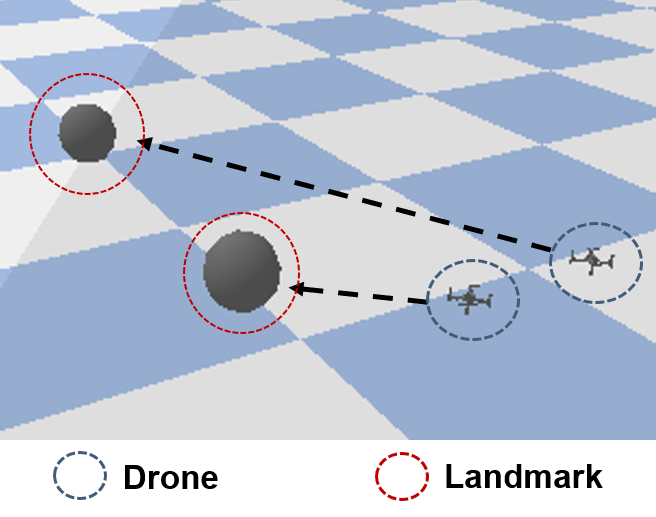}}
    }
	\caption{Experiment environments of Drone.}
\vspace{-6mm}
\label{fig:env_drone}
\end{figure}

To evaluate the effectiveness of our algorithm in large search space, we consider MPE~\cite{mpe} with a massive number of agents and pybullet-gym-drones~\cite{drone} in 3$D$ space as the experimental environments, as shown in Fig.~\ref{fig:env_mpe} and Fig.~\ref{fig:env_drone}. We select three typical tasks from these two environments: Simple Spread, Push Ball, and Drones. Then, we conduct experiments with $N\in\{5, 20, 50\}$ agents in Simple Spread and $N\in\{5, 20\}$ agents in Push Ball in MPE. 
Drone is a more complicated quadrotor navigation task in gym-pybullet-drones, which adopts aerodynamic models of quadrotors to narrow the gap between the simulation and the real world. Therefore, quadrotors in Drone require stronger cooperation to avoid the crash. We conduct experiments with 2 and 4 quadrotors in this 3$D$ simulator. 

\subsubsection{Simple Spread}

\begin{figure*}[ht!]
\captionsetup{justification=centering}
	\centering
    \subfigure[Simple Spread]
	{\centering
        {\includegraphics[height=0.18\textwidth]{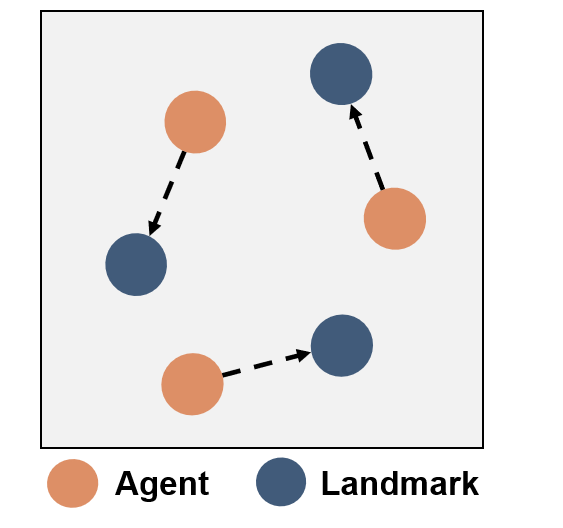}
    	}
    }
    \subfigure[Simple Spread with 20, 50 agents]
	{\centering
         {\includegraphics[height=0.18\textwidth]{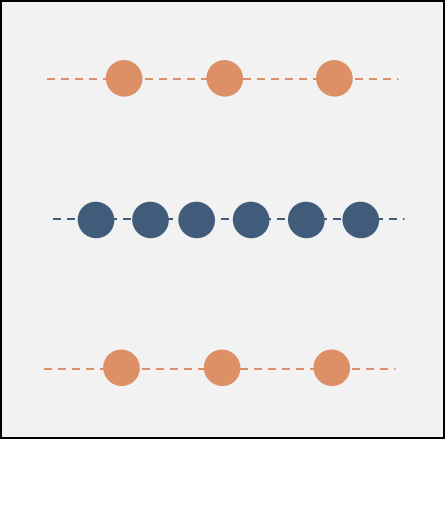}
    	}
    	{\includegraphics[height=0.18\textwidth]{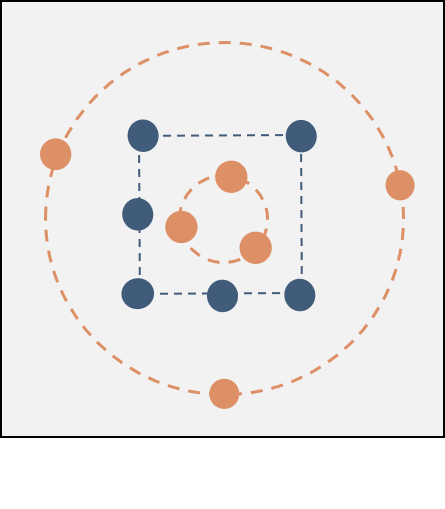}
    	}
    	{\includegraphics[height=0.18\textwidth]{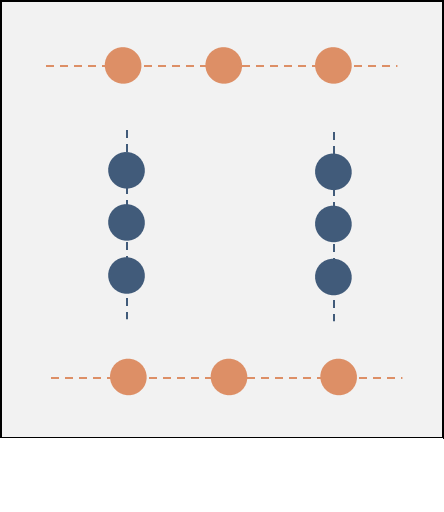}
    	}
    	{\includegraphics[height=0.18\textwidth]{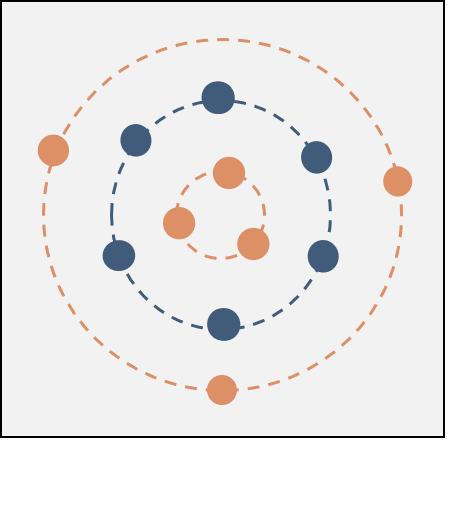}
    	}
    }
    \subfigure[Push Ball]
	{\centering
        {\includegraphics[height=0.18\textwidth]{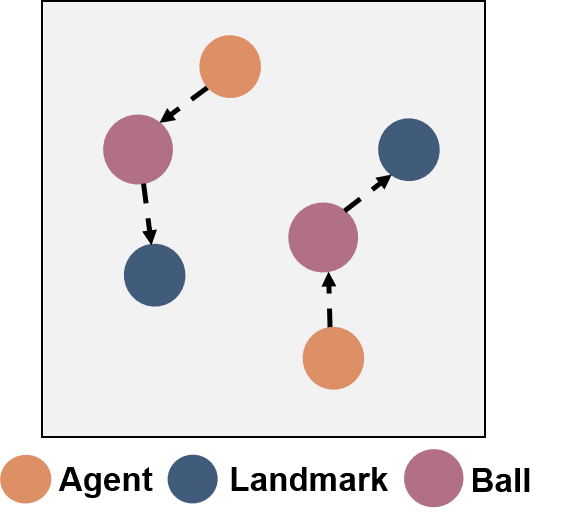}
    	}
    }
    \subfigure[Push Ball with 20 agents]
	{\centering
        {\includegraphics[height=0.18\textwidth]{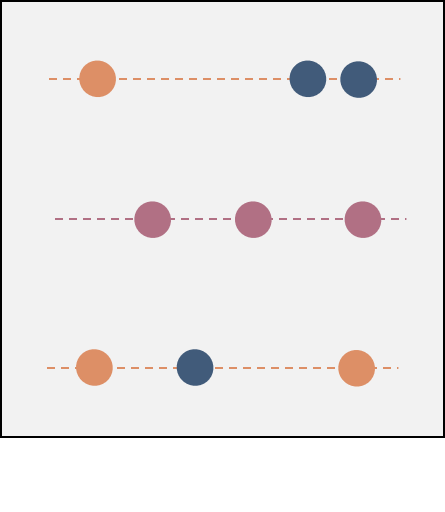}
    	}
    	{\includegraphics[height=0.18\textwidth]{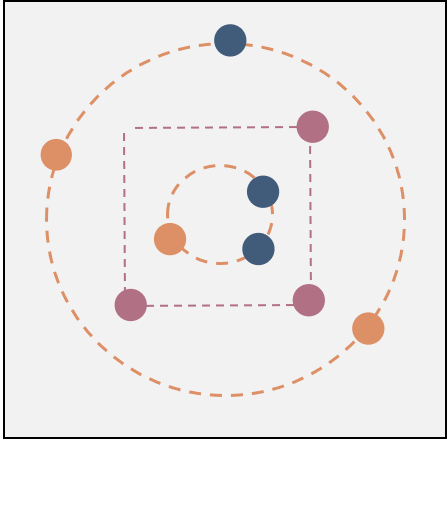}
    	}
    	{\includegraphics[height=0.18\textwidth]{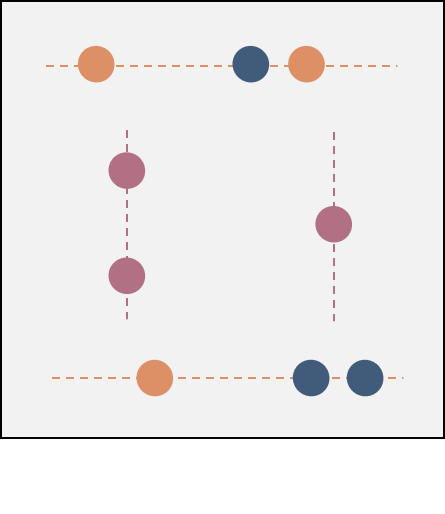}
    	}
    	{\includegraphics[height=0.18\textwidth]{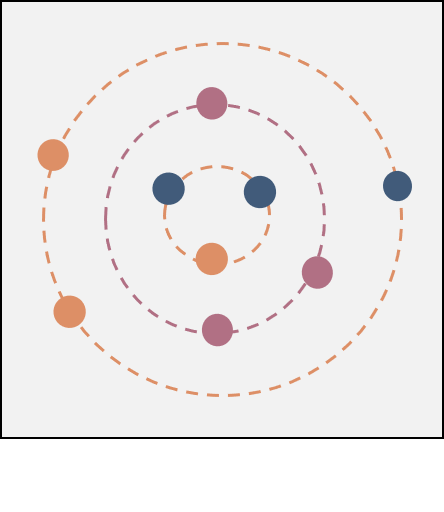}
    	}
    }
	\caption{\raggedright{Experimental environments of Simple Spread and Push Ball in MPE. We remark that (b) represents simplified demonstration of 4 challenging task modes in Simple Spread, where the spawn locations of agents and landmarks are randomly distributed in the orange lines and the blue lines, respectively. (d) expresses simplified demonstration of 4 typical task modes in Push Ball, where the spawn locations of agents and landmarks are randomly distributed in the orange lines, and the spawn locations of balls are randomly distributed in the pink lines.}}
\label{fig:env_mpe}
\end{figure*}

We utilize Simple Spread environment in MPE~\cite{mpe}, a classical 2D navigation task. The episode starts with the initialization of $N$ agents and $N$ landmarks. When all the agents reach the landmarks, this task is 100\% successful. The consequence of the collision between agents is that they will bounce off each other, which is detrimental to navigation efficiency. The available discrete actions of agents include \emph{Up}, \emph{Down}, \emph{Left}, and \emph{Right}. The task's difficulty increases with a larger agent number and map space. The experiment is conducted on 5, 20, and 50 agents. 
In the 5-agent setting, the spawn locations of agents and landmarks are random on the map with a size of 4. 
In the setting of $N\in\{20, 50\}$, the spawn locations of agents and landmarks are randomly initialized in four challenging maps with the size of 36 and 100, respectively, as shown in Fig~\ref{fig:env_mpe}(b). The horizons of steps in $N=5$, $20$, $50$ are 60, 100, and 120, respectively. 

\subsubsection{Push Ball}
Push Ball is a more complicated two-stage navigation task in MPE~\cite{mpe}, requiring each agent to find the ball first, then push it into the landmark. At the beginning of each episode, $N$ agents, $N$ balls, and $N$ landmarks are initialized in the environment. The success rate of the task is the number of agents reaching the landmarks with the balls to the total number of all agents. The available discrete actions of agents include \emph{Up}, \emph{Down}, \emph{Left}, and \emph{Right}.
We conduct the experiments in the setting of 5 and 20 agents. In the scenarios of $N=5$, the spawn locations of agents, balls, and landmarks are randomly distributed on the map of size 16. For $N=20$, their initialization follows four challenging space arrangements on the map with a size of 144, as displayed in Fig~\ref{fig:env_mpe}(d). The horizons of the steps in $N=5$ and $20$ are 100 and 200, respectively. 

\subsubsection{Drone}
We further adopt Drone in pybullet-gym-drones~\cite{drone}, a 3D simulator for flying quadrotors based on pybullet. Drone models the dynamics of quadrotors and controls them by adjusting the torque. Unlike MPE, the collision between quadrotors will lead to a crash and the failure of the task. In our navigation task, the spawn locations of $N$ quadrotors and $N$ landmark are random in the coordinates $x$ and $y$, but fixed in the coordinate $z$, with $x\in(-1.5,1.5), y\in(-1.5,1.5)$ and $z=1.5$. The task's objective is that quadrotors are required to reach all the landmarks on a limited time budget. The action space of each quadrotor is the direction of the acceleration in $x$, $y$, $z$ coordinate, containing \emph{Forward}, \emph{Backward}, and \emph{Stop} in each coordinate. We consider the experiments with $2$ and $4$ quadrotors. In our setting, the frequency at which the physics engine steps is 120 $HZ$; the max speed is 2 $m/s$, and the acceleration is 5 $m/{s}^2$. In every low-level step, we utilize the predicted acceleration direction to calculate each quadrotor's target velocity, and then the quadrotor executes the action every four physics engine steps. The horizon of the steps in $N=2$ and $4$ is 120. 
 
\subsection{Implementation Details}
Each RL training is performed over 3 random seeds for a fair comparison. Each evaluation score is expressed in the format of "mean (standard deviation)", which is averaged over a total of 300 testing episodes, i.e., 100 episodes per random seed. In addition, we use the success rate, the number of landmarks reached by the agents to the total number of landmarks, to express the performance of each algorithm. In the Drone, we additionally consider the collision rate, the number of crashed quadrotors to the total number of quadrotors, to express the capability of cooperation. Our experimental platform involves a 128-core CPU, 256GB RAM, and an NVIDIA GeForce RTX 3090Ti with 24GB VRAM.
\begin{figure*}[ht!]
\captionsetup{justification=centering}
	\centering

        {\includegraphics[height=0.26\textwidth]{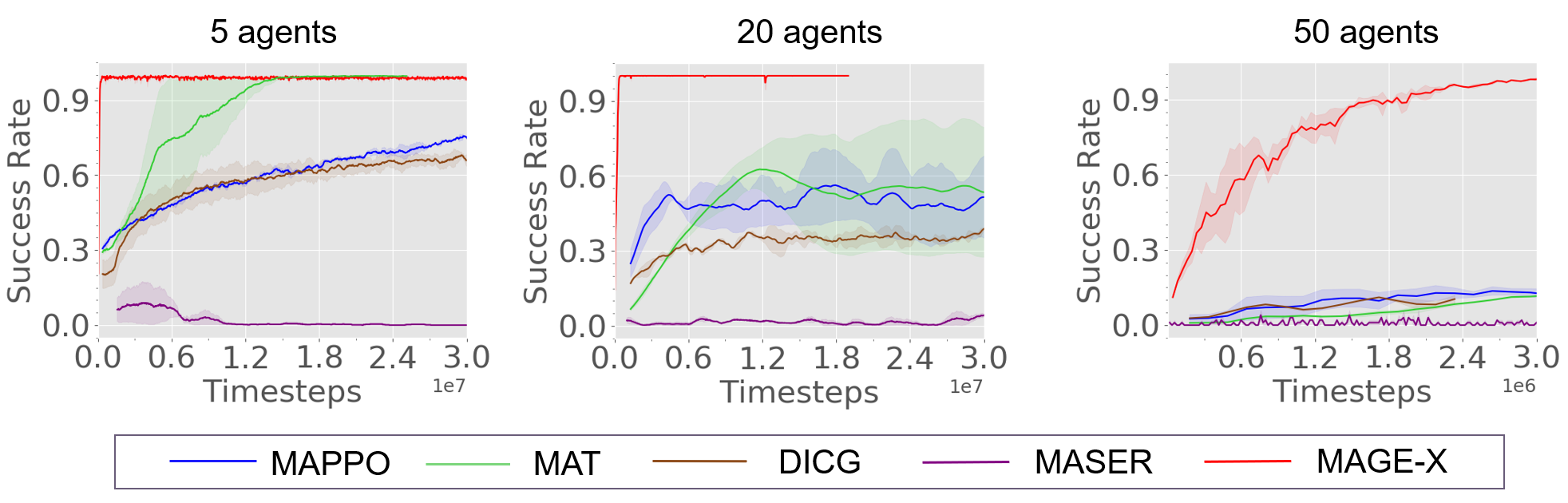}
    	}

	\caption{\centering{Comparison between {\name} and other baselines in Simple Spread with $N=5$, $20$, $50$.}}
\label{fig:baseline_sp}
\vspace{-2mm}
\end{figure*}
\begin{figure}[ht!]
\captionsetup{justification=centering}
	\centering
    {\includegraphics[height=0.24\textwidth]{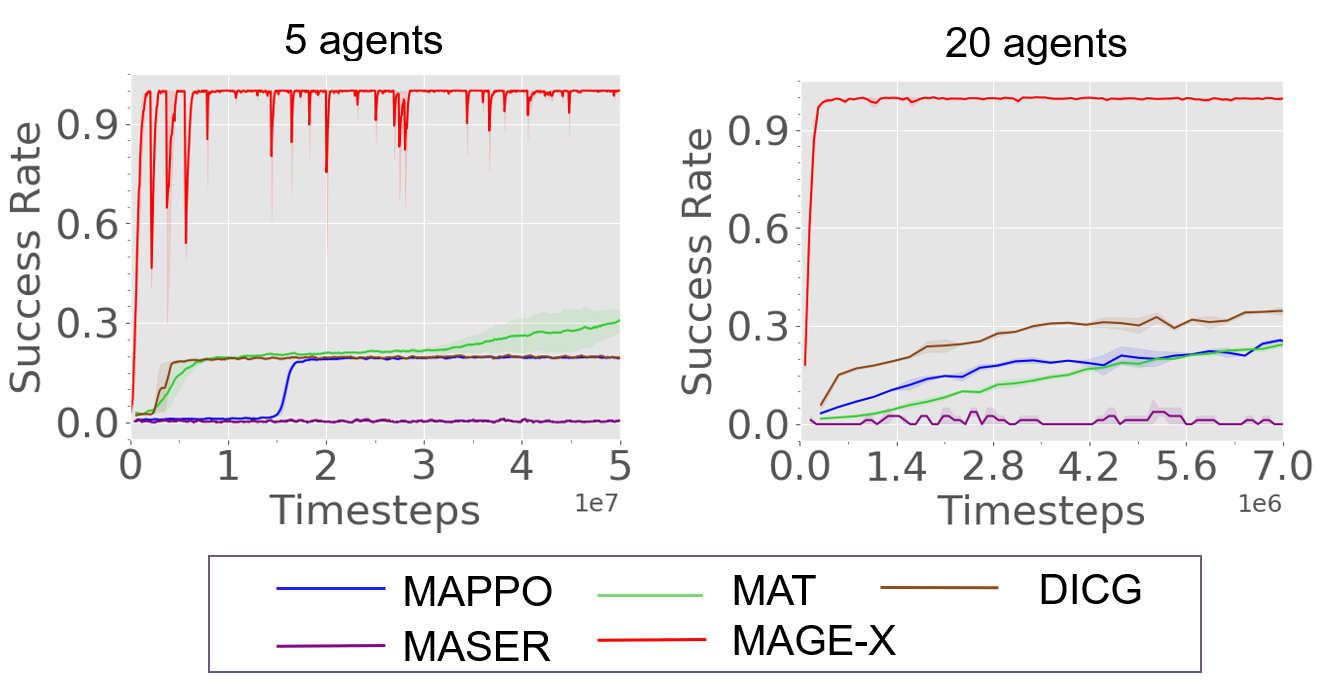}
    }

	\caption{\raggedright{Comparison between {\name} and other baselines in Push Ball with $N=5$, $20$.}}

\label{fig:baseline_pb}
\end{figure}
\begin{figure}[ht!]
\captionsetup{justification=centering}
	\centering     {\includegraphics[height=0.26\textwidth]{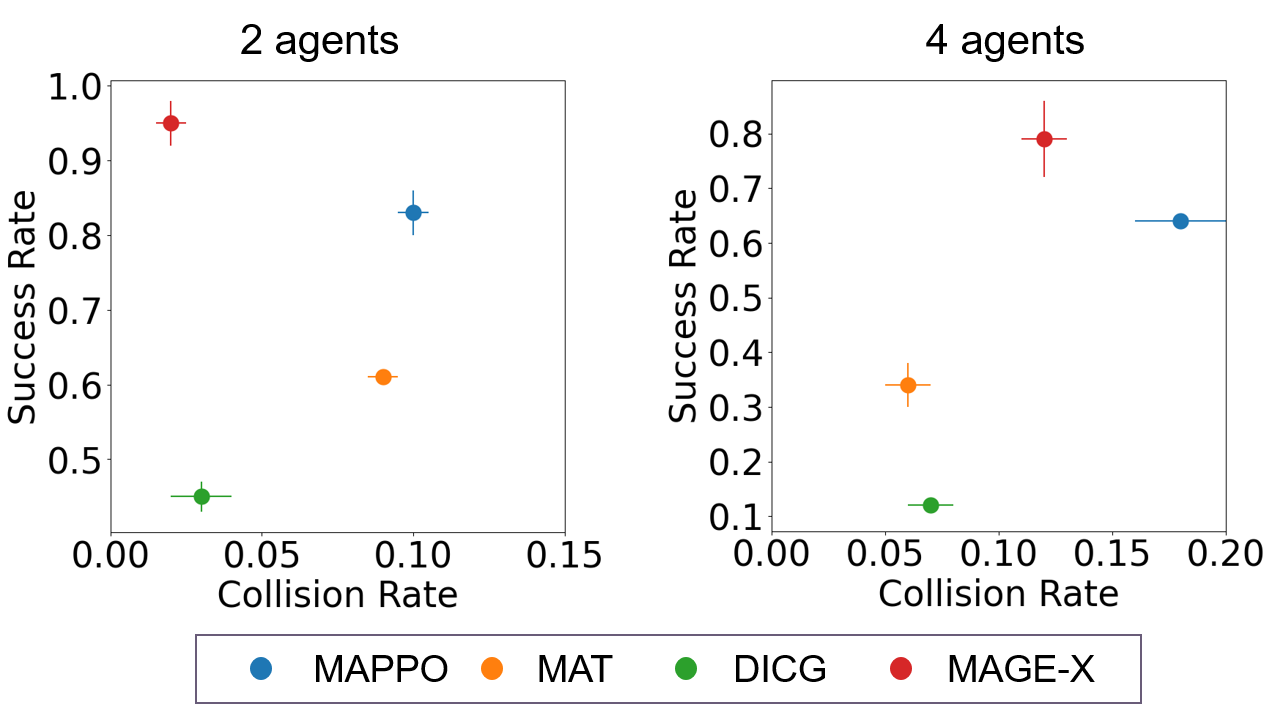}}
	\caption{\raggedright{Evaluation performance of the success rate and the collision rate between {\name} and baselines in Drone. The node in the upper left corner of the figure represents higher performance.}}
\vspace{-5mm}
\label{fig:drone_inter}
\end{figure}

\begin{figure*}[ht!]
\captionsetup{justification=centering}
	\centering
    
    {\includegraphics[height=0.26\textwidth]{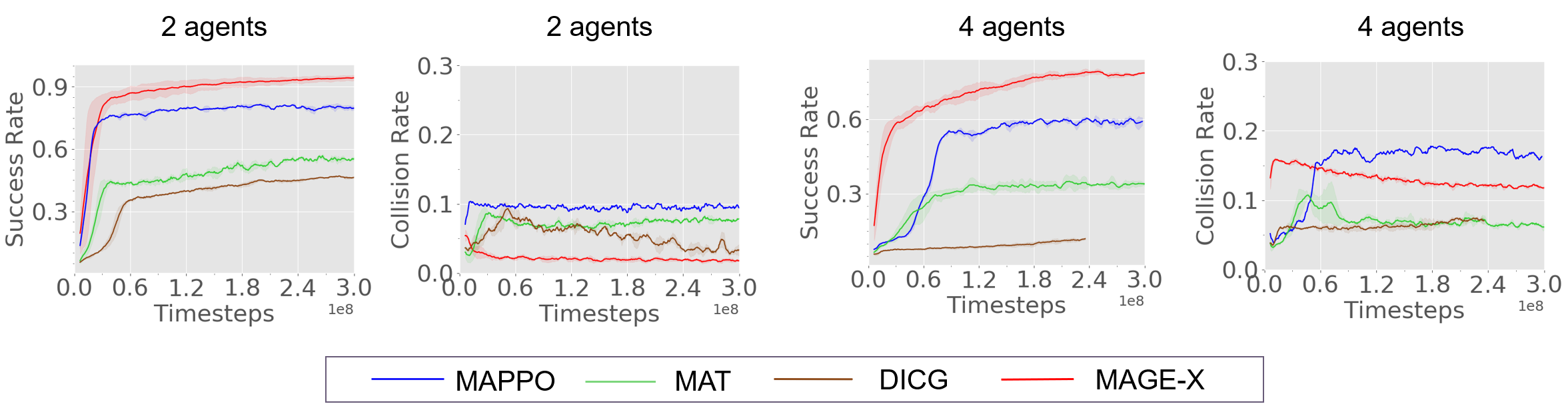}}
     
\vspace{-3mm}
	\caption{\centering{Comparison between {\name} and other baselines in Drone.}}
\label{fig:baseline_drone}

\end{figure*}

\subsection{Baselines}
To demonstrate the effectiveness of our methods, we challenge it against four MARL baseline approaches.
\begin{itemize}
\setlength{\parskip}{0pt} 
\setlength{\itemsep}{0pt plus 1pt}
\item MAPPO~\cite{mappo}: This is the first and the most direct approach for applying PPO in MARL. It equips all agents with one shared set of parameters and updates the shared policy through agents’ aggregated trajectories.
\item Multi-Agent-Transformer~(MAT)~\cite{mat}: This is an encoder-decoder architecture that leverages the multi-agent advantage decomposition theorem to transform the joint policy search problem into a sequential decision-making process.
\item DICG~\cite{DICG}: This graph-based method comprises a module that infers the dynamic coordination graph structure and a GNN module that implicitly reasons about the joint actions or value based on the former module's output.
\item Maser~\cite{maser}: This is a goal-conditioned MARL method that automatically generates subgoals for multiple agents from the experience replay buffer by considering both the individual Q-value and the total Q-value.
\end{itemize}
\subsection{Main Results}
We present and analyze the experiment results of {\name} and MARL baselines in three competitive environments.

\subsubsection{Simple Spread}
We demonstrate the training curves in Fig.~\ref{fig:baseline_sp} and the evaluation performance in \cref{tab: simple_spread}. The results suggest that {\name} performs the best, especially in scenarios with 50 agents, where it converges to the near-optimal solution quickly with a success rate of 100\%, attaining training efficiency more than $10\times$ higher than other competitors. The results show that our algorithm performs the best, especially in the scenario with 50 agents. This graph-enhanced hierarchical framework helps agents quickly handle the problem of reaching targets with high cooperation. MAT excels in other MARL baselines and can achieve a success rate of 100\% in the 5-agent setting, indicating that MAT agents profit from the action information of their teammates. However, as the number of agents increases, the performance of MAT degrades significantly. DiCG is on-par with MAPPO, which indicates that directly utilizing the graph-based method can't contribute to better cooperation and efficiency. Although MASER is also a goal-conditioned HRL method, it is the worst among all the competitors. It implies that the value-based method, MASER, fails to discover effective cooperation strategies within limited training steps since it infers subgoals from the experience replay buffer and lacks a strong connection between agents and target goals.

\begin{table}
\centering

\begin{tabular}{cccccc}
\toprule
  & MAPPO & MAT & DICG & MASER & {\name}  \\ 
\midrule
5 agents  &       0.82\footnotesize{(0.03)}& 1.00\footnotesize{(0.01)}    &  0.71\footnotesize{(0.05)}   &   0.06\footnotesize{(0.12)} & {\bf 1.00}\footnotesize{(0.01)}       \\
\midrule
20 agents &     0.50\footnotesize{(0.31)}  &    0.56\footnotesize{(0.26)} &   0.41\footnotesize{(0.04)}   &  0.02\footnotesize{(0.23)}     & {\bf 1.00}\footnotesize{(0.01)}       \\
\midrule
50 agents &       0.13\footnotesize{(0.02)} &     0.11\footnotesize{(0.01)}&  0.12\footnotesize{(0.02)}    &    0.01\footnotesize{(0.05)}   &    {\bf 1.00}\footnotesize{(0.01)}    \\

\bottomrule
\end{tabular}
\caption{ \raggedright{Evaluation performance of the success rate between {\name} and baselines in Simple Spread.}}
\vspace{-5mm}
\label{tab: simple_spread}
\end{table}

\begin{table}
\centering

\begin{tabular}{cccccc}
\toprule
 & MAPPO & MAT & DICG & MASER & {\name}  \\ 
\midrule
5 agents&     0.20\footnotesize{(0.01)}&  0.32\footnotesize{(0.04)} &  0.20\footnotesize{(0.01)}&  0.01\footnotesize{(0.01)} & {\bf1.00}\footnotesize{(0.01)}        \\
\midrule
20 agents &    0.26\footnotesize{(0.01)}   &  0.25\footnotesize{(0.02)}   &   0.36\footnotesize{(0.01)}   &   0.01\footnotesize{(0.01)}    &     {\bf1.00}\footnotesize{(0.01)}    \\
\bottomrule
\end{tabular}
\caption{ \raggedright{Evaluation performance of the success rate between {\name} and baselines in Push Ball.}}
\vspace{-5mm}
\label{tab: push_ball}
\end{table}





\begin{table}
\centering
\begin{tabular}{ccc|cc}
\toprule
                     & \multicolumn{2}{c|}{2 agents}   & \multicolumn{2}{c}{4 agents}  \\ 
\cmidrule{2-5}  & Suc. Rate & Colli. Rate                                          & Suc. Rate & Colli. Rate       \\ 
\midrule
MAPPO             &          0.83\footnotesize{(0.03)}    &  0.10\footnotesize{(0.01)}&    0.64 \footnotesize{(0.01)}      &   0.18 \footnotesize{(0.02)}               \\ 
\midrule
MAT                 &   0.61\footnotesize{(0.01)}         &  0.09 \footnotesize{(0.01)}                                                      &    0.34\footnotesize{(0.04)}       &   {\bf0.06}\footnotesize{(0.01)}                \\ 
\midrule
DICG  &         0.45\footnotesize{(0.02)}      &     0.03\footnotesize{(0.02)}      & 0.12\footnotesize{(0.01)}&     0.07\footnotesize{(0.01) }                    \\ 
\midrule
{\name}                &          {\bf0.95}\footnotesize{(0.03)}  & {\bf0.02}\footnotesize{(0.01 )}                                            &      {\bf0.79}\footnotesize{(0.07)}        &  0.12   \footnotesize{(0.01)}                  \\
\bottomrule
\end{tabular}
\caption{ \raggedright{Evaluation performance of the success rate and the collision rate between {\name} and baselines in Drones.}}
\vspace{-5mm}
\label{tab: drone}
\end{table}

\subsubsection{Push Ball}
As shown in Fig.~\ref{fig:baseline_pb} and \cref{tab: push_ball}, we conduct the experiments with $N\in\{20, 50\}$ agents in Push Ball. The difficulty of Push Ball lies in that it requires a two-stage goal assignment, i.e., the agents first get the designated ball and then reach the target landmark with the ball. Nevertheless, {\name} still achieves a 100\% success rate with few training timsteps. In contrast, except for MAT, whose success rate slightly increases, other baselines obtain suboptimal policies with a success rate of 20\% in the 5-agent setting. In the scenario with 20 agents, all baselines fail to reach the goals. The results express that the high-level Goal Commander in {\name} has the potential to simultaneously tackle multiple goal assignment problems, for it breaks this $M$\-stage goal assignment into $M$ independent tasks, only requiring the information of assigned targets.\looseness=-1
\subsubsection{Drones}
We report the training performance in Fig.~\ref{fig:baseline_drone} and the evaluation results in Fig.~\ref{fig:drone_inter} and \cref{tab: drone}. The experiment shows that {\name} is superior to other competitors with a success rate of 95\% in $N=2$ and 79\% in $N=4$ in the physically realistic 3D environment, Drone. Unlike MPE, the best competitor in Drone among baselines is MAPPO rather than MAT, indicating that MAPPO has better competence in complex tasks with few agents. Specifically, {\name} outperforms MAPPO, with 12\% higher and 15\% higher success rates in $N=2$ and $4$, respectively. Furthermore, {\name} manifests high coordination with a low collision rate of 0.02\% in $N=2$ and 0.12\% in $N=4$, demonstrating that {\name} succeeds in assigning target goals to agents to coordinate and capture the interaction among agents. Although MAPPO performs best among baselines, its collision rate is remarkably high. We speculate that MAPPO agent pursues high individual capability instead of cooperation. Furthermore, the sample efficiency of {\name} is $2\times$ higher than that of MAPPO in the 4-agent setting.

\subsection{Ablation Study}
To illustrate the effectiveness of each component of {\name}, we consider 3 variants of our method in Simple Spread with 50 agents:
\begin{itemize}
\setlength{\parskip}{0pt} 
\setlength{\itemsep}{0pt plus 1pt}
\item {\bf {\name} w. RG}: We substitute the Goal Scheduler in the Goal Commander with random sampling without replacement to assign each agent a random target goal.
\item {\bf \name-Atten}: We remove GCN in the Obs. Encoder and replace the Graph Encoder in the Action Executor with the attention module~\cite{ATTENTION} to extract the relationship of agents. The concatenation of $(f_o(o_1),f_o(o_2),..f_o(o_N))$ is fed into the attention module.
\item {\bf \name-MLP}: We consider the MLP layer as the alternative to the Action Executor to capture the correlation between agents and goals. The MLP layer takes the concatenation of $(f_o(o_1),f_o(o_2),..f_o(o_N), P_g)$ as input.
\end{itemize}

\begin{figure}
\captionsetup{justification=centering}
	\centering
 {\includegraphics[height=0.20\textwidth]{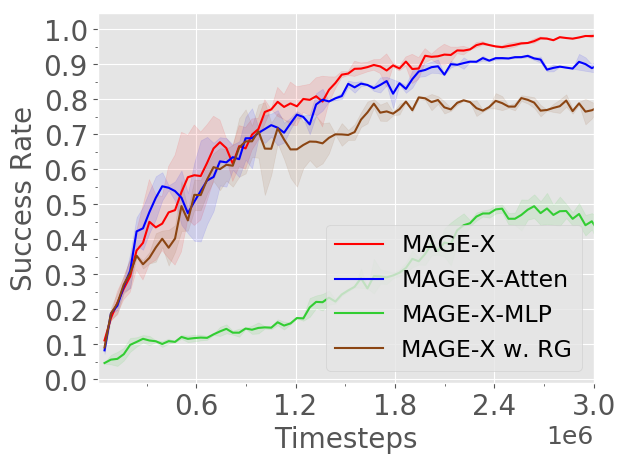}}
	\caption{\raggedright{Ablation study on {\name} in Simple Spread with $50$ agents.}}
\label{fig:ab}

\end{figure}

Fig.~\ref{fig:ab} and \cref{tab: ab} summarize the performance of {\name} and its variants on training and evaluation, respectively. {\name} excels in data efficiency and final performance with a 100\% success rate. \emph{{\name-MLP}} degrades most, implying that the MLP layer is incapable of distinguishing the correlation of agents and target goals from the given information. \emph{{{\name} w. RG}} lacks an appropriate Goal scheduler in the high-level Goal Commander, which may lead to agents being assigned distant goals. Therefore, {{\name} w. RG} reveals a worse performance with an 83\% success rate. \emph{{\name-Atten}} is slightly inferior to {\name} with a 7\% lower success rate. We hypothesize that \emph{{\name-Atten}} provides each agent with the attention weights of all the neighbors, where needless teammates may influence agents. On the contrary, {\name} agent only concentrates on crucial neighbors in the subgraph with useful information.\looseness=-1



\begin{table}
\centering

\begin{tabular}{ccccc}
\toprule
             & \footnotesize{{\name} w. RG} & \footnotesize{\name-Atten} & \footnotesize{\name-MLP} & \footnotesize{\name}  \\ 
\midrule
Suc. Rate &    0.83\footnotesize{(0.01)}   &  0.93\footnotesize{(0.01)}     &   0.50\footnotesize{(0.08)}    &    {\bf1.00}\footnotesize{(0.01)}    \\

\bottomrule
\end{tabular}
\caption{ \raggedright{Evaluation performance of Success Rate between {\name} and its variants in Simple Spread.}}
\label{tab: ab}
\vspace{-9mm}
\end{table}
\section{Conclusion and Future Work}
In this paper, we propose a goal-conditioned MARL framework, \emph{Multi-Agent Commander-Executor}~(\name), to improve data efficiency and cooperation in multi-agent navigation tasks, especially in scenarios with large space spaces~(e.g., a massive number of agents or complex 3$D$ simulator). {\name} consists of a high-level Goal Commander and a low-level Action Executor, where the Commander allocates target goals to agents via the probability distribution of goals and the Executor leverages GNN and a Goal Encoder to capture team representation with strong goal guidance. Thorough experiments demonstrate that {\name} achieves higher sample efficiency and better performances than all the state-of-the-art MARL baselines in multi-agent particle environments~(MPE) with large-scale agents and a more complicated quadrotor navigation task. Currently, {\name} mainly focuses on multi-agent navigation tasks, and we will try to apply {\name} to other multi-agent tasks beyond navigation in the future. \looseness=-1

\section*{ACKNOWLEDGMENT}

This research was supported by National Natural Science Foundation of China (No.62203257, U19B2019, M-0248), Tsinghua University Initiative Scientific Research Program, Tsinghua-Meituan Joint Institute for Digital Life, Beijing National Research Center for Information Science, Technology (BNRist) and Beijing Innovation Center for Future Chips.

\clearpage
\setcounter{section}{0}
\renewcommand\thesection{\Alph{section}}
We would suggest to visit \url{https://sites.google.com/view/mage-x23} for more information.

\section{{\name} Details}
\subsection{Hungarian algorithm}
The Hungarian algorithm~\cite{Hungarian} is a combinatorial optimization algorithm which solves the assignment problem. To be specific, the assignment problem can be formulated with a bipartite graph. Assume that we have a complete bipartite graph $G=(S,T;E)$ with $n$ worker vertices $S$ and $n$ job vertices $T$, and each edge has a non-negative cost $c(i,j)$. We want to find a perfect matching with a minimum total cost. And the graph and cost can also be translated as a matrix, where the element in the $i_{th}$ row and $j_{th}$ column represents the cost of assigning the $j_{th}$ job to the $i_{th}$ worker. 
However, the Hungarian algorithm is not the exact optimal solution and yields an approximately optimal solution with unaffordable time complexity $O(n^3)$. Therefore, we propose a learning-based module, the Goal Commander, to figure out a (nearly)-optimal solution with less inference computation. And we implement the Hungarian algorithm as an expert to judge the performance of the Goal Commander by calculating the reward formulated as:
    \begin{equation*}
    R_{c} = 1-\frac{C_{c}}{C_{h}}.
    \end{equation*} 
$R_{c}$ represents the distance cost of our assignment strategy, $C_{c}$, against the distance cost of the Hungarian algorithm, $C_{h}$.

\subsection{Observation Space and Action Space}
For the Goal Commander, the observation space is the positions of all agents, $(x, y)\in \mathbb{R}^2, x, y \in[0, map size]$, and the positions of target goals, $(g_x, g_y)\in \mathbb{R}^2, x, y \in[0, map size]$. The action space of the Goal Commander is the index of assigned goals for agents, $A_{com} =\{i, i\in N\} $, where $i$ is the index of assigned goals, and $N$ represents the number of goals. 
For each Action Executor $i$, the observation space contains agent $i$'s position $(x_i, y_i)\in \mathbb{R}^2, x_i, y_i \in[0, map size]$, and its target goal was given by the commander $(g_i^x, g_i^y)\in \mathbb{R}^2, g_i^x, g_i^y \in[0, map size]$, other agents' positions $(x_{-i}, y_{-i})\in \mathbb{R}^2, x_{-i}, y_{-i} \in[0, map size]$, and one hot label $L_i$ to represent whether it has reached the goal. The action space of each Action Executor is the primitive action for each agent, $A_{exe}^i =\{a^i|a^i \in action\}$. We denote that $action$ is the available primitive action, which is a discrete variable sampled from the categorical distribution. 
\subsection{Action Generator}
The Action Generator takes in the team features with strong goal guidance from the State Extractor then infers a primitive action for each agent to complete the navigation tasks. 

\setcounter{table}{0}
\begin{table}[h]
\centering
\begin{tabular}{cc}
\toprule
common hyperparameters      & value  \\
\midrule
gradient clip norm          & 10.0 \\
GAE lambda                   & 0.95    \\              
gamma                      & 0.99 \\
value loss                        & huber loss \\
huber delta                 & 10.0   \\
mini batch size           & batch size {/} mini-batch  \\
optimizer                & Adam       \\
optimizer epsilon            & 1e-5       \\
weight decay             & 0          \\
network initialization  & Orthogonal \\
use reward normalization   &True \\
use feature normalization   &True \\
learning rate & 2.5e-5 \\
parallel environment threads & 10 \\
number of local steps & 15\\
\bottomrule
\end{tabular}
\vspace{2mm}
\caption{MAPPO Hyperparameters}
\label{tab:hyperparameter}
\end{table}

\begin{table}[h]
\centering
\begin{tabular}{cc}
\toprule
architecture hyperparameters      & value  \\
\midrule
$f_{sche}$          & $[4\times N, 1\times N,1]$ \\
$f_{o}$                   & $[2\times N, 32,1]$  \\      
$f_{g}$                      & $[32\times N, 32\times N,1]$ \\
$f_{goal}$                        & $[2\times (N+1), 32, 1]$ \\
$f_{state}$                        & $[32\times 2, 32, 1]$ \\
feature size                 & 32   \\
$l_{g}$                      & 2     \\
\bottomrule
\end{tabular}
\vspace{2mm}
\caption{Architecture Hyperparameters}
\label{tab:architecture}
\end{table}
\section{Experiments}
\subsection{Training Details}
We train our policy with MAPPO~\cite{mappo}, which is a multi-agent extension of PPO~\cite{ppo}. The training hyperparameters can be found in \cref{tab:hyperparameter} and the detailed architecture of \name\ is shown in \cref{tab:architecture}.

In {\name}, we implement feature normalization and reward normalization to achieve more robustness. We use Adam optimizer with optimizer epsilon as $1e^{-5}$, and the learning rate is $2.5e^{-5}$. Value loss is Huber loss with Huber delta $10.0$. The gradient clip norm we choose is $10.0$ and GAE lambda is $0.95$. Discounted factor $gamma$ of the expected reward is $0.99$, and no weight decay has been used. Parallel threads of the environment are $10$ and the number of local steps is $15$. For neural network architecture, it is orthogonal initialized with feature size $32$.  
\subsection{Baseline}
In {\name}, we compare against one planning-based competitor, Multi-Agent $A*$, and four MARL baseline algorithms, MAPPO, MAT, DICG, and Maser. 
\subsubsection{Multi-Agent A*~(MA A*)~\cite{astar}} MA A* is the Multi-Agent version of $A*$ algorithm. It first transforms the continuous space into grid space and calculates the cost of each grid. The cost is the distance between the grid and the agent, and the distance between the grid and the goal. Besides, it also considers the collision among agents. The target is to find a path from the agent to the target goal with the minimal total cost.
\subsubsection{MAPPO~\cite{mappo}}
MAPPO, the Multi-Agent version of PPO, achieves surprisingly strong performance in StarCraftII, Google Research Football, MPE, and Hanabi. Compare with common value-based Multi-Agent algorithm~\cite{qmix, qtran, maddpg}, Mappo is the first and the most direct, approach for applying PPO in MARL. It equips all agents with one shared set of parameters and updates the shared policy through agents’ aggregated trajectories, and achieves competitive or superior results in final rewards without losing sample efficiency.  
\subsubsection{MAT~\cite{mat}}
MAT is an encoder-decoder architecture which leverages the multi-agent advantage decomposition theorem to transform the joint policy search problem into a sequential decision-making process. Unlike prior arts such as Decision Transformer fit only pre-collected offline data, MAT is trained by online trials and errors from the environment in an on-policy fashion.
\subsubsection{DICG~\cite{DICG}}
DICG is a graph-based method consisting of a module that infers the dynamic coordination graph structure, and a GNN module that implicitly reasons about the joint actions or value based on the former module's output. It allows the agent to learn the trade oﬀ between full centralization and decentralization via standard actor-critic methods to significantly improve coordination for domains with large numbers of agents.
\subsubsection{Maser~\cite{maser}}
Maser is a goal-conditioned MARL method that automatically generates subgoals for multiple agents from the experience replay buffer by considering both the individual Q-value and the total Q-value. It adopts the individual intrinsic reward for each agent to reach the assigned subgoals and maximize the joint action value.

\subsection{Additional Results}
\subsubsection{Compared with MA A*}
\cref{tab: simple_spread} shows the performance of {\name} and the planning-based baseline, MA A*, in Simple Spread. It expresses that MA A* has a comparable performance with {\name} in $N=$ 5, 20 agents, but shows a worse result in $N=$ 50. It implies that MA A* has the potential to handle the multi-agent multi-target problem, but it may be difficult to find optimal paths when the number of agents increases.

\begin{table}
\centering

\begin{tabular}{cccc}
\toprule
 Agents & 5 &20 & 50  \\ 
\midrule
MA A*   &    {\bf 1.00}\footnotesize{(0.01)}& 0.93\footnotesize{(0.03)}  &   0.77\footnotesize{(0.03)}  \\
\midrule
{\name}  &     {\bf 1.00}\footnotesize{(0.01)}  &    {\bf 1.00}\footnotesize{(0.01)} &   {\bf 1.00}\footnotesize{(0.01)}\\ 

\bottomrule
\end{tabular}
\caption{ \raggedright{Evaluation performance of the success rate between {\name} and MA A$^*$ in Simple Spread.}}
\label{tab: simple_spread_all}
\end{table}

\clearpage
\bibliographystyle{ACM-Reference-Format} 
\bibliography{main}


\end{document}